\documentclass[conference]{IEEEtran}
\IEEEoverridecommandlockouts
\usepackage{cite}
\usepackage{amsmath,amssymb,amsfonts}
\usepackage{algorithm}
\usepackage{algorithmicx}
\usepackage[noend]{algpseudocode}
\usepackage{lipsum}

\usepackage{graphicx}
\usepackage{enumerate}
\usepackage{textcomp}

\usepackage{xcolor}
\usepackage{subcaption}
\usepackage[normalem]{ulem}
\usepackage{comment}
\usepackage{booktabs}
\usepackage{array}
\usepackage{tabularx}
\usepackage{longtable}
\usepackage{threeparttable}

\algnewcommand{\LeftComment}[1]{\Statex \(\triangleright\) #1}

\def\BibTeX{{\rm B\kern-.05em{\sc i\kern-.025em b}\kern-.08em
    T\kern-.1667em\lower.7ex\hbox{E}\kern-.125emX}}

\usepackage{tikz}

 \newcommand\copyrighttext{%
   \footnotesize “This work has been submitted to the IEEE for possible publication. \\
   Copyright may be transferred without notice, after which this version may no longer be accessible.”}

 \newcommand\copyrightnotice{%
     \begin{tikzpicture}[remember picture,overlay]
         \node[anchor=south,yshift=10pt] at (current page.south) {\fbox{\parbox{\dimexpr\textwidth-\fboxsep-\fboxrule\relax}{\copyrighttext}}};
     \end{tikzpicture}%
 }

\begin{document}
\copyrightnotice
\captionsetup[table]{skip=0pt} 

\title{EON-1: A Brain-Inspired Processor for Near-Sensor Extreme Edge Online Feature Extraction

}


\author{
\IEEEauthorblockN{
Alexandra Dobrita\IEEEauthorrefmark{1}$^,$\IEEEauthorrefmark{3}, 
Amirreza Yousefzadeh\IEEEauthorrefmark{2},
Simon Thorpe\IEEEauthorrefmark{4},
Kanishkan Vadivel\IEEEauthorrefmark{1},
Paul Detterer\IEEEauthorrefmark{1}}
\IEEEauthorblockN{
Guangzhi Tang\IEEEauthorrefmark{5}, 
Gert-Jan van Schaik\IEEEauthorrefmark{1}, 
Mario Konijnenburg\IEEEauthorrefmark{1},
Anteneh Gebregiorgis\IEEEauthorrefmark{3}}
\IEEEauthorblockN{
Said Hamdioui\IEEEauthorrefmark{3},
Manolis Sifalakis\IEEEauthorrefmark{1}
}
\IEEEauthorblockA{\textit{\IEEEauthorrefmark{1}imec, Netherlands}, \textit{\IEEEauthorrefmark{2}University of Twente, Netherlands}, 
\textit{\IEEEauthorrefmark{3}Technical University of Delft, Netherlands}, }
\IEEEauthorblockA{ \textit{\IEEEauthorrefmark{5}Maastricht University, Netherlands},
\textit{\IEEEauthorrefmark{4}}CNRS, France}
}

\maketitle

\begin{abstract}


For Edge AI applications, deploying online learning and adaptation on resource-constrained embedded devices can deal with fast sensor-generated streams of data in changing environments. However, since maintaining low-latency and power-efficient inference is paramount at the Edge, online learning and adaptation on the device should impose minimal additional overhead for inference. 

With this goal in mind, we explore energy-efficient learning and adaptation on-device for streaming-data Edge AI applications using Spiking Neural Networks (SNNs), which follow the principles of brain-inspired computing, such as high-parallelism, neuron co-located memory and compute, and event-driven processing. We propose EON-1, a brain-inspired processor for near-sensor extreme edge online feature extraction, that integrates a fast online learning and adaptation algorithm. We report results of only 1\% energy overhead for learning, by far the lowest overhead when compared to other SoTA solutions, while attaining comparable inference accuracy. Furthermore, we demonstrate that EON-1 is up for the challenge of low-latency processing of HD and UHD streaming video in real-time, with learning enabled.

\end{abstract}


\section{Introduction}

With the emergence of Edge AI applications, enabling online learning and adaptation on resource-constrained embedded devices is becoming increasingly appealing, as it has the potential to tackle a wide range of challenges: first, it can deal with on-the-fly adaptation to fast sensor-generated streams of data under changing environments \cite{snn-ol}. Second, it could facilitate learning from limited amounts of training points and third, it can alleviate a variety of hindering factors associated with offline training in the Cloud \cite{CHEN2020264}, such as incurred energy consumption of sensor data transfers and extra memory storage for the training samples, but also data privacy and security concerns. Addressing these challenges, however, would be of little benefit if the cost of enabling on-device learning and adaptation, in terms of energy overhead for inference, is high.

For low-latency and low-power inference, the main genres of neural models under consideration currently include Quantized and Binary Neural Networks (QNNs/BNNs) \cite{qnn, bnn}, as well as their neuromorphic recurrent siblings, Spiking Neural Networks (SNNs) \cite{snn}.
For these models, training offline with back-propagation \cite{bp} (and more recently feedback alignment \cite{feedback}) have been well established for achieving high performance. However, when it comes to online on-device learning/adaptation, there is no satisfactory solution to date that can provide both good performance and resource efficiency. On the one hand, ``vanilla'' back-propagation and several of its variations are not scalable and highly resource demanding for edge device deployments, due to the need for global signalling, high-precision data and end-to-end state tracking that grows exponentially with network depth \cite{b1,b2}. On the other hand, local learning schemes, such as Spike Timing Dependent Plasticity (STDP) \cite{shouval2010stdp} and BCM \cite{bcm} among others, although resource efficient, are lagging behind in performance, even for networks of mediocre depth. In between these two extremes of the spectrum, a few other gradient-based methods have been proposed \cite{spoon, eprop}, but have not gained mainstream traction until now because they have not yet achieved a satisfactory balance between resource efficiency and performance. As a result, the topic remains subject of active research, with a small handful of custom processors having emerged, that embed some sort of limited online learning or adaptation capability \cite{tbcas2109}, \cite{odin},  \cite{tcsii2023}, \cite{b7}, \cite{tcsi2022}. The key differentiators among them is typically energy, area efficiency, speed of learning, and the supported learning rule.

In this paper we explore the merits of the process of Ultra-Rapid Visual Categorization (URVC) \cite{b8} in the mammalian visual cortex, as a paradigm for fast online on-device learning and adaptation at the Edge. In URVC, fast analysis and classification of images is attributed to efficient encoding and transmission of information from the retinal ganglion cells to the orientation-selective cells in the visual cortex in the shortest time possible (i.e., with the first emitted spike) via the optic nerve \cite{b9, b10}. Past work has defended the efficiency of URVC for few-shots learning (2 to 5 input samples) \cite{posterThorpe} with binary STDP \cite{b11} in networks of integrate-and-fire (IF) neurons for tasks such as face identification \cite{b12} and natural scene recognition \cite{b13}, \cite{b14}, \cite{b15}.

Building upon these past ideas, we introduce EON-1, an energy- and latency-efficient custom Edge processor for Spiking and Convolutional Neural Networks, that uses 1-bit synaptic weights and 1-spike per neuron. EON-1 embeds a very fast on-device online learning algorithm (amortizable for few-shot learning) inspired by URVC and the works discussed above, which, when benchmarked in an ASIC node, achieves less than 1\% energy overhead for on-device learning (relative to the inference). To our knowledge, our solution incurs the least energy overhead for learning on-device, compared to state-of-the-art solutions, showing a better efficiency by at least a factor of 10x. We report that EON-1 consumes 0.29mJ to 5.97mJ for converging to accuracies between 87.8\% - 92.8\% on the MNIST classification task, from randomly initialized weights for both inference and on-device learning. With an energy consumption of 148.4nJ - 663.5nJ per processed sample, only 1.2nJ/sample is used for learning. We extend our solution to a practical use case of feature extraction on UHD frames (8M pixels per frame), demonstrating the scalability of the proposed hardware architecture towards real-time processing, in parallel with learning.

In the remaining of this paper, Section II presents the main learning and inference algorithm used in this work, Section III presents the hardware architecture of EON-1, section IV presents and discusses the benchmarking results, while Section V presents the main conclusions of our work.

\section{Algorithm and refinements}
\label{sec:algo}

\subsection{Overview of ultra-rapid feature extraction in the visual cortex }

In the biological visual system, it has been observed that ultra-rapid, high-level feature extraction occurs in the orientation-selective first layer of the visual cortex based on the \emph{first spike} emitted by ganglion cells in the retina \cite{b16}. In order to generate the first spike that carries sufficient information for high-level recognition, the  ganglion cells in the retina act as intensity-to-latency converters \cite{b17}, such that they emit spikes in the order of the strength of the visual stimulation. 

This behavior is leveraged by a hardware-efficient spike encoding scheme and can be modelled in a Convolutional Neural Network (CNN) with orientation-selective edge filters in the first layer \cite{b18}, followed by a temporal 1-winner-take-all (1-WTA) circuit that laterally inhibits the orientation cells \cite{b11}, such that only the dominant edge filters are allowed to propagate information downstream (channel-wise max-pooling).

\subsection{Methods}
The hardware-optimized algorithm underlying the inference and learning in EON-1 is partially inspired by \cite{b14}, \cite{b15} and \cite{b19}. It comprises of two functionally separate parts: a fixed, input part consisting of convolutional filters that are pre-trained and relevant to the input data modality (i.e., sensor domain), and a volatile (latent) part consisting of one or more layers of integrate-and-fire (IF) neurons, that are subject to learning and adaptation. The main insight here is that, typically, as long as the input data modality does not change, the filters in the early layers will not or do not need to get modified upon retraining or fine-tuning~\cite{yosinski2014transferable, he2016resnet}. Instead, what is often practised is that, when the application task changes, only the latter (near the output) layers of a network need to be modified upon training or fine-tuning. A second insight derives from the theoretical equivalence \cite{b21}, \cite{b22}, \cite{b23} between arbitrary deep versus wide neural networks (with up to two hidden layers). This means that from a hardware efficiency point of view, we can opt for an architecture that in the second part has fewer very-wide layers (thousands to millions of neurons) instead of many (deep) narrow layers, which favors hardware parallelization, low-latency and learning simplicity with a local rule. In the extreme case of EON-1, we have used only a single IF neurons layer (that plays a role similar to an associative memory).

Following, we explain the inference and learning processes in the proposed algorithm. Although we explain inference first, it's important to note that, similarly to the biological brain, learning and inference will run concurrently in the proposed hardware architecture of EON-1 (Section \ref{sec:hw_arch}). Throughout the rest of the paper we use the nomenclature in Table \ref{tab:nomenclature} to describe our work. 
\begin{table}[htb]
\begin{center}
    \caption{Nomenclature used for algorithm description}
    \label{tab:nomenclature}
    \begin{tabular}{|p{0.25\linewidth} | p{0.65\linewidth}|}
        \hline
        \textbf{Name}   & \textbf{Description}  \\
        \hline
        $D$ & Spike and weights vectors are $D \times D$ in size \\
        \hline
        $K\_S$ & Convolution filters are $K\_S \times K\_S$ in size. For convolution with a stride of 1 and no zero-padding, input images are of size $(D + K\_S - 1)^2$ \\
        \hline
        \textit{active} weight & Synaptic weight fully ON ($1'b1$)\\
        \hline
        \textit{inactive} weight & Synaptic weight fully OFF ($1'b0$) \\
        \hline
        
        \textit{ineffective} spike & Input spikes residing at inactive synaptic weights positions\\
        \hline
        \textit{ineffective} weight & Active synaptic weight not contributing to pattern recognition \\

        \hline   
    \end{tabular}
\end{center}
\end{table}

\begin{figure}[t]
    \centering
    \includegraphics[width=\columnwidth]{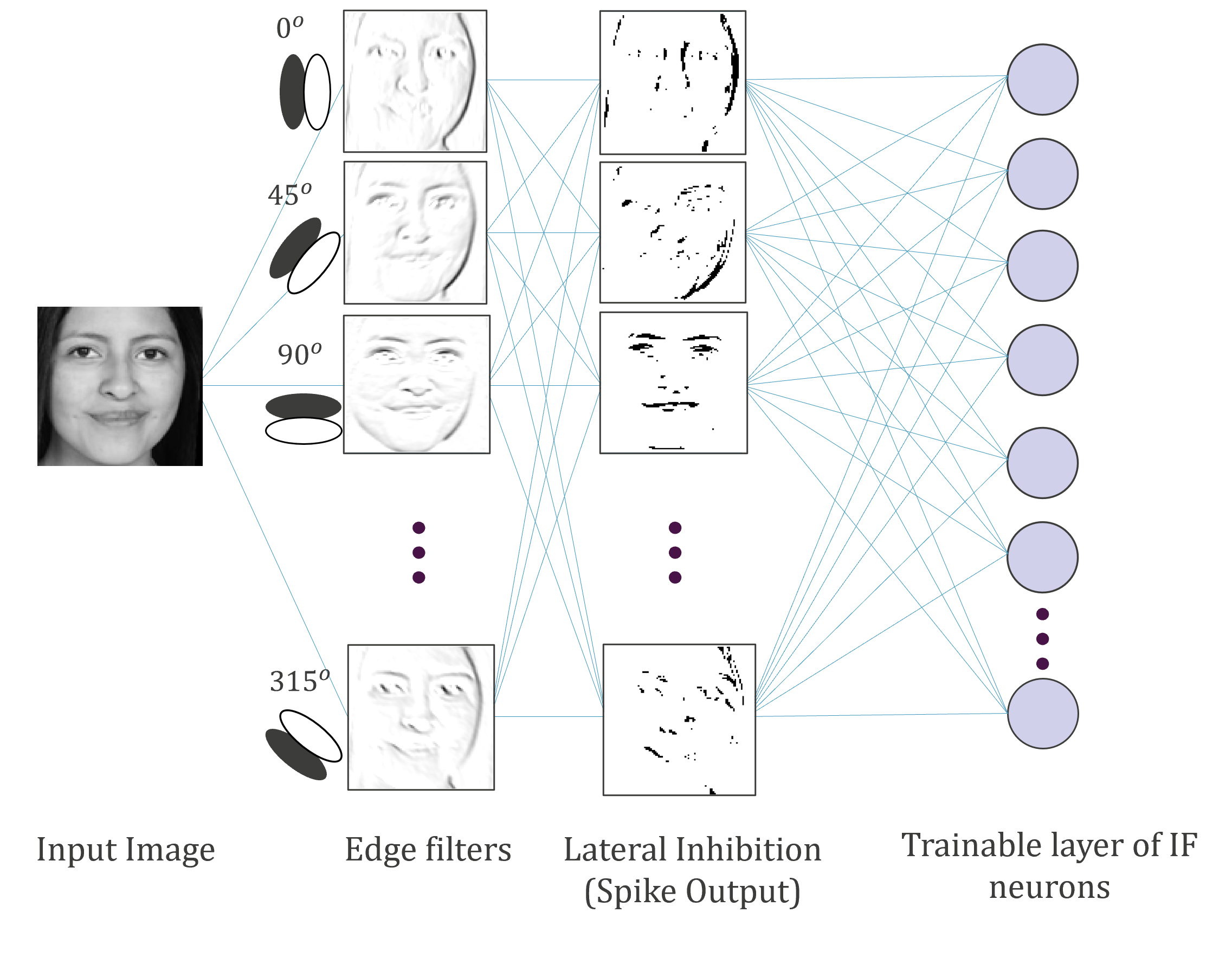}
    \caption{The neural network structure for this work. It includes a layer of edge-filtering convolution, a lateral inhibition layer, and a layer of fully connected neurons equipped with binary STDP training.}
    \label{fig:algo_inference}
    
\end{figure}


\subsubsection{Low-latency, 1-spike--based inference}
\label{sec:algo_inference}

\begin{figure}[ht]
    \centering
    \includegraphics[width=0.55\columnwidth]{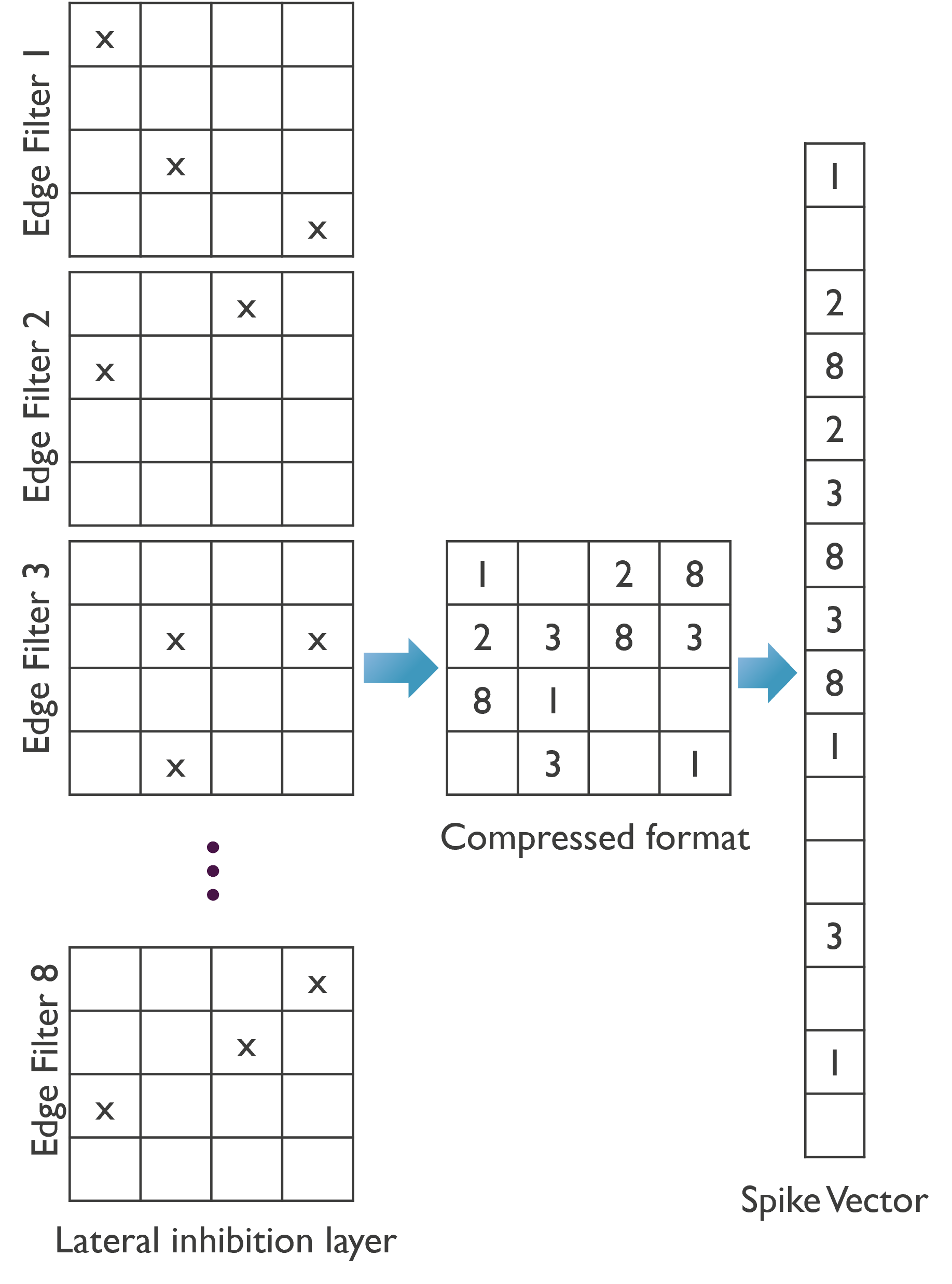}
    \caption{The lateral inhibition layer produces binary spikes that can be compressed into a spike vector. Each element of the resulting vector designates the index of the source edge filter that generated the first spike for the respective pixel position.}
    \label{fig:spike_vector}
\end{figure}

The network structure is depicted in Fig.\ref{fig:algo_inference}. It starts with processing input images of size $(D + K\_S - 1)^2$ with (pre-defined and domain-specific) edge filters of size $K\_S^2$ in the first layer, followed by a lateral inhibition layer. The latter step ensures that for every $(X, Y)$ location, only one spike is fired from all the edge filtering orientation cells (i.e., the winner edge filter is selected through channel-wise max-pooling). The result is a highly sparse binary spiking output with a special structure that allows it to be encoded (compressed) in a $D \times D$ spike vector representation, as shown in Fig. \ref{fig:spike_vector}. 

\begin{figure}[!h]
    \centering
    \includegraphics[width=0.8\columnwidth]{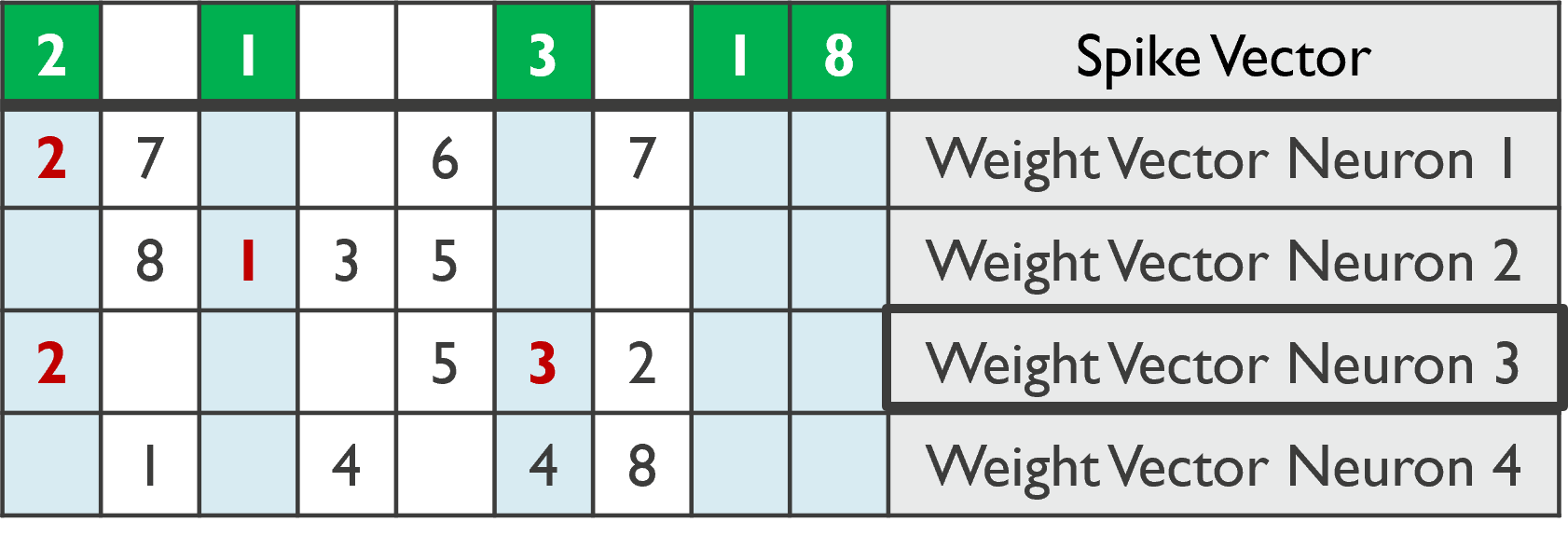}
    \caption{Example of a spike vector and four weight vectors. Bold elements in the weight vectors are the one that match the spike vector. Assuming $T_{Fire} = 2$, only neuron 3 fires.}
    \label{fig:weight_vectors}
\end{figure}

Subsequently, the output of the lateral inhibition layer is fully connected to a layer of IF neurons, where each neuron's synaptic weights meet the following constraints:

\begin{itemize}    
     \item The synaptic weights are binary, which means they either fully connect (activate) or disconnect (deactivate) synapses.
     \item All neurons have the exact same amount of active synapses ($W$ parameter in Table \ref{tab:neuron_params}).
     \item For every neuron, the synaptic connection to only one edge filter is active at each $(X, Y)$ location in the input image.
\end{itemize}
These constraints enforce very sparse synaptic weights vectors, which can be encoded in the same compressed format as the spike vector (see example in Fig. \ref{fig:weight_vectors} for $W=4$), rendering our algorithm to be a hardware-friendly option for neuromorphic processing systems. Note that weight sparsity is inherent in our algorithm due to the constraints we apply (parameter $W$), rather than a result of the training process. Consequently, the network's connections are sparse right from the start and remain sparse, even after learning.
\begin{table}[htb]
\begin{center}
    \caption{Learning hyper-parameters for the IF Neurons layer}
    \label{tab:neuron_params}
    \begin{tabular}{|p{0.2\linewidth} | p{0.7\linewidth}|}
        \hline
        \textbf{Name}   & \textbf{Description}  \\
        \hline
        $N$        & Number of IF neurons in learning layer \\
        \hline
        $W$             & Number of active synaptic connections per neuron \\
        \hline
        $T_{Learn}[0]$  & Initial learning threshold \\
        \hline
        $T_{Fire}$      & Firing threshold. Can be fixed, or variable (fraction of $T_{Learn}$, but $\infty$ at the beginning).  \\
        \hline  
        $swap\_rate$    &  Fraction of ineffective synapses that are concentrate  (swapped) on active but ineffective spikes (input lines), during learning. \\
        \hline  
        $K$             & Maximum number of neurons which are allowed to learn (an input spike vector), such that $K << N$\\
        \hline   
    \end{tabular}
\end{center}
\end{table}

Finally, when elements in the weight vector of an IF neuron downstream from the lateral inhibition layer align in index position (pixel position) and value (edge filter index) with corresponding elements of a spike vector, the respective IF neuron integrates the incoming spikes and performs a threshold comparison. Thereafter, the neuron fires if the number of matching elements exceeds the firing threshold, $T_{Fire}$.

\subsubsection{Fast online learning (few-shots)}
\label{sec:algo_learning}
During inference, not all of the $W$ active synaptic weights of the IF neurons downstream from the lateral inhibition layer contribute to pattern recognition (i.e., to $T_{Fire}$ being reached), hence, we denote them by \textit{ineffective} synaptic weights. Likewise, we denote as \textit{ineffective} the spikes in the input spike vector that do not align in position with any active synaptic weights.
However, to increase selectivity to input patterns, the synaptic weights can be trained by swapping ineffective synaptic connections to ineffective spike positions.

The training rule that we use is a variant of STDP, namely stochastic binary STDP \cite{b20}, and is triggered independently for each neuron. By contrast to gradient-based learning with a global signal (like in back-propagation), here the choice of a local rule: (a) is functionally justified by the shallow nature of the network architecture (i.e., a global signal is not required to traverse a shallow model structure); (b) has the advantage that inherently it learns in an unsupervised manner (i.e., it does not need to rely on labels); and c) is architecturally well suited for efficient hardware implementation given its parallelization potential and infrequent application on each neuron independently: each IF neuron can learn independently of all other, when the learning threshold ($T_{Learn}$ in Table~\ref{tab:neuron_params}) is exceeded. Subsequently, after each update, this threshold increases gradually, to reduce the neuron plasticity and increase its selectivity to a specific pattern. Optionally, if forgetting needs to be incorporated in the algorithm for out-dated patterns and non-stationary input distributions, the learning threshold may be set to slowly decay.   

Another feature of our STDP rule is its binary synaptic weights. In conventional STDP, where the real-valued weights for synapses without a pre-synaptic firing would gradually reduce (Long Term Depression, LTD) and the weights for synapses with a pre-synaptic activity would gradually increase (Long Term Potentiation, LTP), a neuron can slowly adapt to new data patterns without forgetting the already learned patterns. By contrast here, with the use of binary weights, it is impossible to adapt the weight values gradually. Nevertheless, by increasing/decreasing the maximum number of neurons $K$ that are allowed to update their weights during learning and the maximum number of ineffective synapses per neuron that can be swapped during an update step ($swap\_rate$), we can control the learning rate (or equivalently forgetting rate) even under such an extremely low weight precision regime.

\begin{algorithm}[H]
\caption{Binary stochastic STDP for $K = 1$ } \label{alg:stdp}
\begin{algorithmic}
\State \textbf{Input data:} A $D^2$-bit spike vector, $s$, and swapping rate, $swap\_rate$.
\For{each randomly chosen IF neuron}
\State $\triangleright$ Read its synaptic weights vector, $w$, and its $T\_learn$ and update its membrane potential, $V\_mem$
\State $V\_mem = \sum_{i = 0}^{D^2 - 1} s(i) \land w(i) $

\State $\triangleright$ Evaluate learning condition

\If{$V\_mem \geq T\_learn$} \Comment{Update weights}

\State $swap\_N = swap\_rate \times (W - V\_mem)$
        \For{$s = 0$ to $swap\_N - 1$}
            \State $ineff\_s =  \text{find } ((s \land (\neg w)) = 1)$ 
            
            \State $ineff\_w = \text{find }((\neg s \land w) = 1)$  

            \State $rand\_ON\_idx = \text{randperm }(ineff\_s, 1)$
            \State $rand\_OFF\_idx = \text{randperm }(ineff\_w, 1)$

            \State $w[rand\_ON\_idx]  = 1'b1$
            \State $w[rand\_OFF\_idx] = 1'b0$

        \EndFor
        \State $T\_learn = T\_learn + swap\_N$
        \State \textbf{break} \Comment{Exit process once $K = 1 $ neuron has learned}
    \EndIf
\EndFor
\State \textbf{return} $w$, $T\_learn$
\end{algorithmic}
\end{algorithm}

\begin{figure}[ht]
    \centering
    \includegraphics[width=0.8\columnwidth]{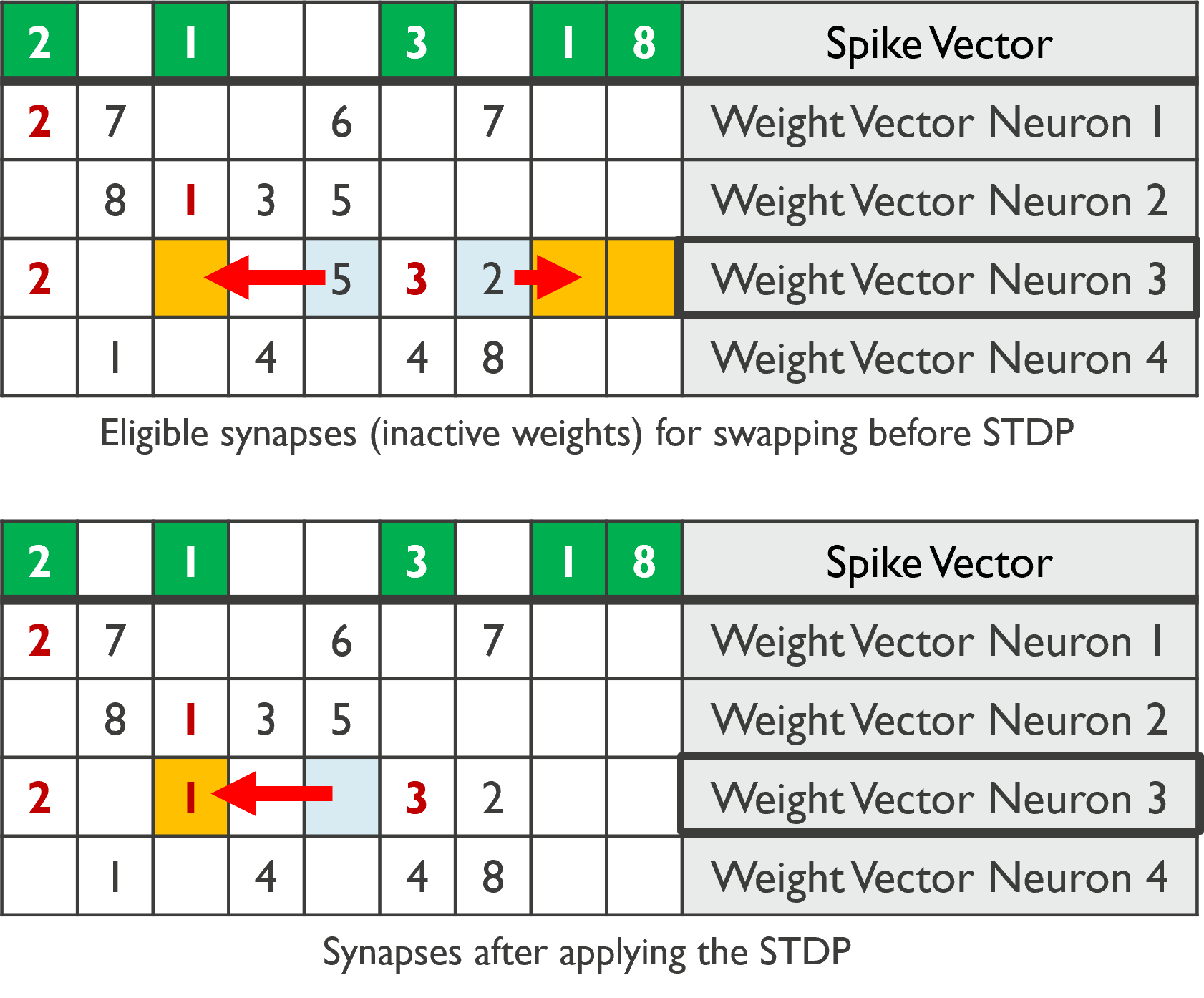}
    \caption{An example of our stochastic binary STDP algorithm. Top: neuron $3$ is eligible for learning and it has two ineffective synapses (blue). Ineffective spikes positions are highlighted in orange. Bottom: one ineffective weight is randomly selected and swapped with the ineffective spike. }
    \label{fig:stdp_weight_vectors}
\end{figure}

Finally, learning progresses according to the steps described in Algorithm  \ref{alg:stdp} and exemplified in Fig. \ref{fig:stdp_weight_vectors}:

\begin{enumerate}[i.]

\item
\textbf{Perform inference and select neurons to learn}: For an incoming spike vector, update the membrane potential of each randomly chosen neuron and evaluate if learning condition is true. If true, trigger the learning process, otherwise proceed to evaluating the next neuron. 
In Fig.\ref{fig:stdp_weight_vectors} (top), the membrane potentials of neurons 1-4 will then be $1$, $1$, $2$ and $0$ respectively. Assuming $T_{Learn}=2$, in Fig. \ref{fig:stdp_weight_vectors} (top), neuron $3$ will be eligible to update its weights.

\item
\textbf{Update (bit-flip) weights}: This step starts with evaluating how many ineffective weights will be swapped ($swap\_N$ in algorithm \ref{alg:stdp}), followed by finding the vector positions at which ineffective spikes reside. Subsequently, weights are swapped as follows: a weight is activated at a random ineffective spike position, while another weight is deactivated at a random ineffective weight position. Fig. \ref{fig:stdp_weight_vectors} illustrates this mechanism for neuron $3$. As a consequence of learning, next time a variant of the same input pattern appears, neuron 3 will have higher probability of firing. 

\item
\textbf{Increase learning threshold $T_{Learn}$} by an amount that equals the number of synapses swapped. This mechanism is inspired by the homeostasis observed in biological neurons \cite{yin2015structural} for selectivity towards frequently seen patterns. Thereafter, the firing threshold $T_{Fire}$, may be a function of the learning threshold (thus being adaptive) or can also be fixed and independent. In our implementation we choose the former option.

\end{enumerate}
    
\subsubsection{Classifier}
%
For multinomial classification tasks, a low overhead readout classifier can be configured by clustering together groups of IF neurons (from the trainable layer) for each class. The class label then serves as a supervision signal to select the respective cluster of neurons that will be allowed to learn the stimulus.
During inference, the predicted class is derived through cluster voting, by selecting the cluster that exhibits the highest firing/activation rate in its neurons.

\begin{figure}[ht]
    \centering
    \includegraphics[width=\columnwidth]{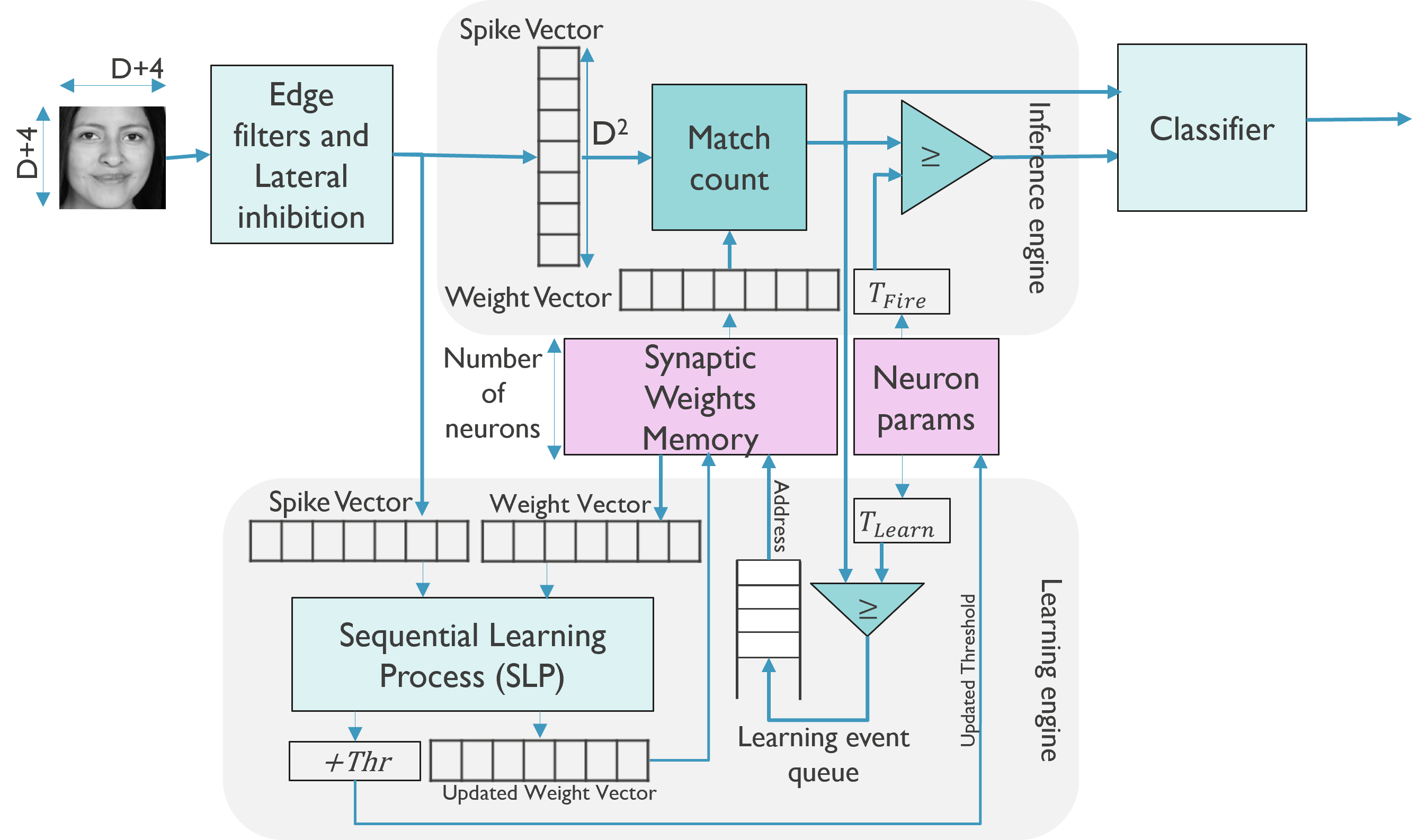}
    \caption{Base hardware architecture of the EON-1 processor, equipped with an inference engine and a learning engine which embeds the proposed STDP-based learning rule. This architecture can be flexibly and trivially scaled-up by vectorizing either of: the IF units, the edge-filter units, and/or the sequential-learning processes (depending on design-space requirements)}
    \label{fig:hardware}
\end{figure}
\begin{figure}[ht]
    \centering
    \includegraphics[width=0.5\columnwidth]{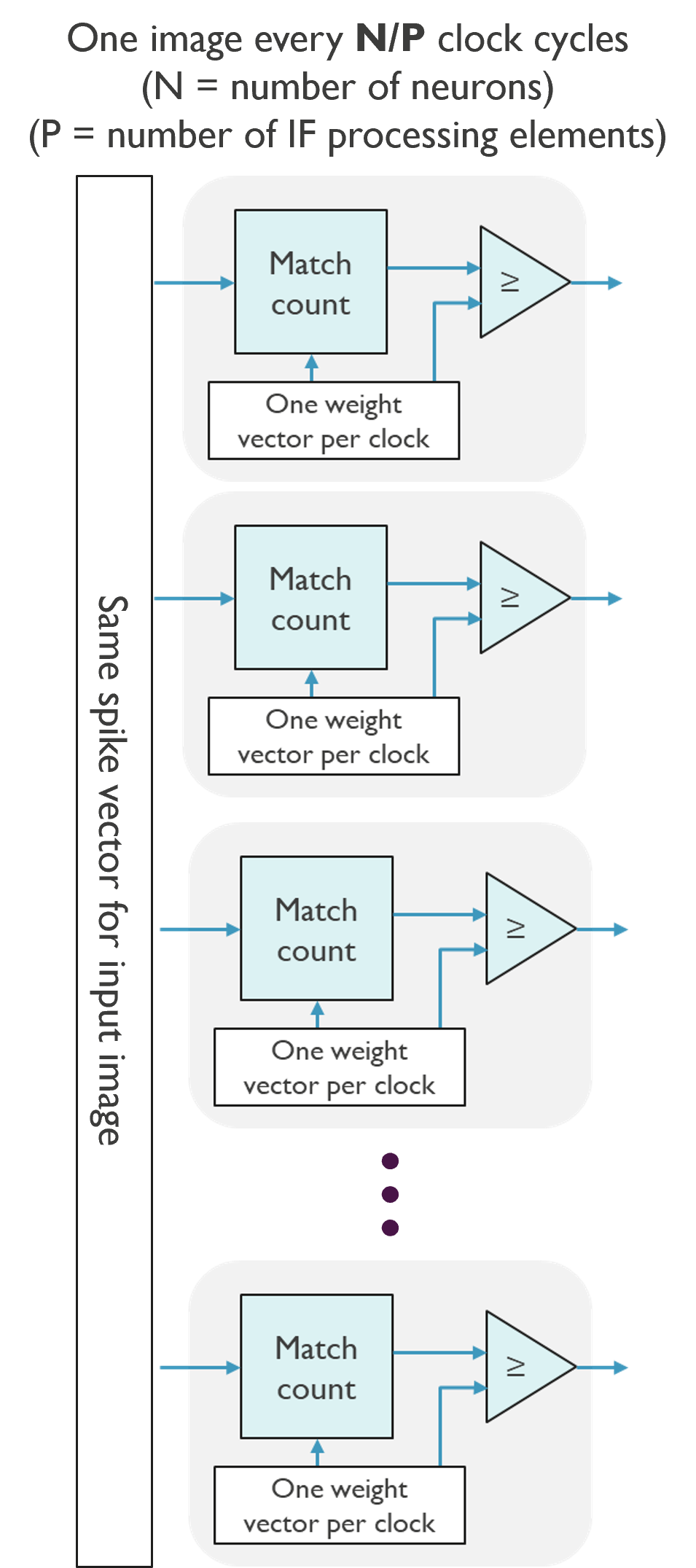}
    \caption{To increase parallelism during inference with very-wide output layers, we can spawn several IF processing elements in a P-wide vector pipeline. Each IF processing element can update one neuron per clock cycle, such that for $N$ neurons in the system, each input image takes $N/P$ clock cycles.}
    \label{fig:multiple_match_units}
\end{figure}
\section{Hardware architecture}
\label{sec:hw_arch}
 The hardware architecture of the EON-1 neural processor is depicted in Fig. \ref{fig:hardware}. Notice that the inference and learning parts are intertwined, highlighting the fact that on-device learning is not an add-on but rather an embedded feature of the processor. Execution-wise, the two engines perform concurrently. Moreover, in this work, both spike encoding and classifier run on-chip, in parallel with learning and inference. Following, we describe the architectural details of EON-1's digital hardware implementation.


\subsection{Edge-Filters and Lateral-Inhibition} This block implements the first two layers of the proposed network in Fig. \ref{fig:algo_inference}, receiving as input image pixels streamed row-by-row every clock cycle and producing a spike vector as output. Internally, this is achieved through a shift-register that buffers $K\_S$ rows. Once the shift register is fully loaded, all edge filters are applied in parallel, resulting in a throughput of one output row per clock cycle. To illustrate, for $K\_S = 5$, and an input image of size  $(D + 4)^2$ (i.e., convolution with stride 1 and no zero-padding), a spike vector of $D^2$ size will be ready in $D$ clock cycles. For $8$ edge filters (in this work) we need $4$ bits for each compressed element in the spike vector, to represent the $9$ possibilities (zero indicates the \emph{no-firing} case). The spike-encoding is pipelined with the inference engine and therefore the overhead of loading the shift register is hidden after the first encoded spike vector. Moreover, to accommodate various application performance requirements and bandwidths of the input pixels but also hardware platforms, this block is fully parametrisable, such that a trade-off between area and latency can be made, i.e., throughput and latency can be customised.  

\subsection{Synaptic Weights Memory and Neuron Parameters Memory} These blocks store the weight vectors and parameters of the IF neurons in the trainable layer. In this work, the learning hyper-parameters (Table \ref{tab:neuron_params}) are shared among all neurons and thus only the learning and firing thresholds need to be stored. The learning and inference engine have shared access to these memory blocks. 

\subsection{Inference engine} This engine is based on a match counter unit (i.e., it sums the positional matches in the spike and weight vectors) and a firing threshold comparator, which essentially implement the operation of an IF neuron. The IF operation executes in one clock cycle, while all neurons in the trainable layer are updated in a time-multiplexed manner. However, the inference engine can be accelerated by enhancing the system with multiple IF units that can operate in parallel, as in Fig. \ref{fig:multiple_match_units}: for a layer of $N$ IF neurons, and provided $P$ IF processing units are available, the inference can execute in $N/P$ clock cycles.

\subsection{Learning engine}
\label{sec:hw_learning}
The bottom half of the hardware architecture shown in Fig. \ref{fig:hardware} contains the building blocks of implementing the on-device \textit{event-driven} online learning functionality. Typically, learning takes place slower than inference, but this can be partly controlled by the maximum number of neurons $K$ (Table \ref{tab:neuron_params}) allowed to learn at a time. The learning engine has the following components:
\begin{itemize}
    \item \textit{Learning-threshold Comparator and Learning Event Queue}: During inference, in addition to comparing an IF neuron state (output of the match count block) against the firing threshold, it is also compared against the learning threshold. If this is exceeded, the address of the corresponding neuron is pushed to the learning event queue. The active capacity of this queue limits the maximum number of neurons that can undergo learning triggered by one input sample (image) and is thus set by the parameter $K$ mentioned earlier. 
    \item \textit{Sequential Learning Process (SLP)} is a FSM that consumes the learning event queue. At each clock cycle, SLP examines one position of a weight vector. If it encounters an ineffective synapse, it swaps it with an ineffective spike location. The order in which the weight vector elements are inspected is randomly generated using an LFSR unit. The sequential learning process stops after a sufficient number of swaps has been reached ($swap\_N$ in algorithm \ref{alg:stdp}).
    Since learning is a slower process than inference, sequential rather than concurrent processing is a more area-efficient solution. The exact time it takes for each weight vector to be updated depends on the weight vector dimension ($D^2$) and the number of swap operations (worst-case number of swaps is $W - T\_learn[0] $). For a worst-case scenario, where a weight vector update takes always $D^2$ clock-cycles, the learning process for $N$ neurons takes $K \times D^2$. To ensure that the learning process never lags behind inference, $K \times D^2$ should remain smaller than $N$. Multiple SLPs can be instantiated in parallel, if necessary (similar to IF neuron units).
\end{itemize}
Note that since only up to $K << N$ neurons will be scheduled-in for learning every input spike vector, a question of bias strategy arises, as to which neurons should be prioritised. Plausible strategies can be to select the top-$K$ least-recently fired neurons (i.e., least recently seen pattern), or highest membrane state (most likely pattern). However, these are costly solutions that require sorting and incur latency cost. Selecting the first-$K$ is by far the fastest and cheapest solution, yet erratic (likely, only a small subset of neurons will participate in learning) if the neurons are always sequenced in the same order. Thus, to ensure that all IF neurons have the same fair chance of being scheduled-in for learning, during inference a random number generator produces a different start address every time for sequencing the IF neurons, thus providing an inexpensive arbitration.
\subsection{Classifier circuit} This module is based on $log_2^W$-bit adders and provides the output prediction after inference based on the index of the cluster with the highest activity, i.e., it computes the maximum firing activation per class. This circuit is pipelined with the IF units, and thus provides its result a clock cycle after the last updated neuron. 

\section{Results and discussion}

\subsection{Experimental setup}

For measuring EON-1's performance, we evaluated two instantiations of it, one on FPGA, using a Xilinx VU37P HBM chip at 100MHz frequency, and one using gate-level post-synthesis ASIC simulation\footnote{Cadence Genus\cite{genus} for synthesis and Cadence JOULES\cite{joules} for time-based power analysis} at 500 MHz frequency, for a typical corner of $0.8V$ and $25^{\circ}C$ in a GF-22nm node.

The reported results are based on two tasks: the first task is multiclass classification based on MNIST \cite{mnist} which, despite its simplicity, allowed us to benchmark against other state-of-the-art solutions from a hardware-efficiency point of view, both for FPGA and ASIC instantiations. For this task, to minimize area and power, we downscale the MNIST samples to $14\times14$. However, we note that our algorithm is resilient to inputs downscaling (i.e., it keeps good performance for smaller resolutions, benefiting hardware resources). The second task is a binary class face detection based on faces \cite{utkface} and CIFAR-10\cite{cifar} datasets, which enabled us to explore the unique features of our algorithm: one is online adaptation starting from a pre-trained network, while for the second we evaluate the suitability of EON-1 for real-time processing of streaming high-definition (UHD) data. We present results that validate both the algorithm effectiveness for online learning/adaptation and the hardware efficiency for on-device learning (primary goal).  

\begin{threeparttable}[htb]
    \caption{Comparison of EON-1 with other FPGA solutions benchmarked on MNIST}
\label{tab:fpga_benchm}
    \footnotesize
    \setlength\tabcolsep{0pt}
\begin{tabular*}{\linewidth}{@{\extracolsep{\fill}} l cc cc c @{}}
    \toprule
     & \textbf{TCSI'21} & \textbf{TCSII'21} & \textbf{Neuro'17} & \textbf{ICTA'23} &\textbf{EON-1} \\
        & \cite{tcsi2021} & \cite{tcsii2021} & \cite{neurocomputing2017} & \cite{icta2023} & this work\\

        \midrule
  Accuracy & 85.28\% &  90.58\% &  89.1\% &  95.49\%   & 87.8\% -- 92.8\% \\
        \midrule
        Neur. model     & LIF & LIF & LIF & LIF & IF \\
        \# Neurons        & 300  &  2304 & 1591 & 320 & 2000 -- 9000 \\
        \# Synapses      & 176800  & NR & 638208 & NR & 1.6M -- 7.2M \\
        Weight Prec.     & 16b float  & 2b & 16b fixed & 8b & 1b \\
        Encoding         & rate  & rate & rate & rate & rank-1 \\
        \midrule
        \multicolumn{6}{c}{Learning specs} \\
        \midrule
        Online  & yes  & yes & yes & NR & yes \\
        On-chip & yes  &  no & yes & yes & yes \\
        Rule    & STDP & STDP & STDP &SG \tnote{2} & SB-STDP \tnote{1} \\
        \midrule
        \multicolumn{6}{c}{Hardware specs} \\
        \midrule
        FPGA Chip        & Virtex-7 & ZCU102 & Virtex-6 & VC707 & VU37P \\    
        Clock Freq.      & 100MHz & 200MHz & 120MHz & 115MHz & 100MHz \\* 
        \midrule
        \multicolumn{6}{c}{Resource utilization} \\
        \midrule
         LUT & NR    & 2209  & 71598   & 22779   & 8053   \\ 
   FF & NR   & 447   & 50905   & 15072   & 1637 \\
   BRAM & NR   & 451  & 204  & NR  & 24 -- 108 \\* 
        \midrule
        \multicolumn{6}{c}{Throughput (fps)}  \\
        \midrule
        Learning & 61 & NR & 0.05 & NR & 10.9K --  47.2K \\
        Inference & 285 & 46.44 & 0.11 & 1183 & 11K -- 49.6K \\
  
        \bottomrule
\end{tabular*}
\begin{tablenotes}[para,flushleft,small]
        \item NR: Not Reported\, 
        \item[1] SB-STDP: Stochastic Binary STDP\,
        \item[2] SG: Surrogate gradient, a spike-friendly variant of back-propagation 
\end{tablenotes}
\end{threeparttable}
\vspace{1.5em}

For consistent algorithm behaviour across all tasks and tests, we kept the same learning hyper-parameters throughout, which were set for very fast, few-shots learning. All accuracy measurements were performed on a testset (different than the training set) with deactivated on-device learning. The common parameters throughout all the tests and tasks are:
\begin{itemize}
    \item 
 
    $swap\_rate = 1$, i.e., all ineffective synaptic weights are swapped to ineffective spikes during learning. Thus, only one training epoch is sufficient to memorize a pattern. Note that, for each neuron, its weight vector won't necessarily match the input spike vector completely since there are typically more ineffective spikes than $W$, and only a random selection of those is matched, thus allowing variability in the learned patterns and preventing overfitting.
 
   \item
   $T{Learn}[0] = 6$ is the initial learning threshold. We chose this value empirically. Since all synaptic
weights are initially random, this value should be high enough to denote a sufficient degree of
coincidence between the input and a prototypical neuron model, such that learning is triggered early-on,
but small enough to ensure that the more salient features in the input are learned.

    \item

    $T_{Fire} = T_{Learn}/2$. Recall that $T_{Fire}[0] = \infty$ to force the inference circuit to be silent before any learning has taken place. After the first learning event for an IF neuron, its $T_{Fire}$ is set to this value, activating it for inference. This value has also been experimentally obtained to give an optimal ratio of time that the inference circuit is active compared to the learning circuit (taking into account that the latter is slower). 
    
    \item   
    $K = 1$ to allow only one neuron, randomly chosen among all the eligible neurons, to adapt its weights every time learning is triggered. For few-shots learning, it is important to avoid wasting model memory when multiple neurons learn and lock in the same pattern, while $K>1$ allows better generalization (lower selectivity) with slower few-shot learning ($swap\_rate < 1$). 

        \item   
    $W = 64$ active connection are allowed during all experiments.  
\end{itemize}


\begin{table*}
\centering
\begin{threeparttable}

\caption{Comparison of EON-1 with other ASIC solutions benchmarked on MNIST}
\label{tab:asic_benchm}
    \footnotesize
    \setlength\tabcolsep{0pt}
\begin{tabular*}{0.8\linewidth}{@{\extracolsep{\fill}} l cc cc cc @{}}
    \toprule
        & \textbf{TCSI'22} & \textbf{TBCAS'19} & \textbf{TBCAS'19} & \textbf{TCSII'23} & \textbf{ISCAS'20} & \textbf{EON-1} \\
        & \cite{tcsi2022} & \cite{tbcas2109} & \cite{odin} & \cite{tcsii2023} & \cite{spoon} & this work \\
        \midrule
        

        Accuracy & 93\% &  87.4\% &  84.5\% &  93.54 \%   & 92.8\% - 95.3\%\tnote{$\ddag$} & 87.8\% -- 92.8\%  \\
        \# Weight updates to accuracy  \tnote{$\triangleq$}& 160k & 60k & NR &  60k & 60k & 2k -- 9k \\
        Learning cost for accuracy & 65.5mJ & 23.73mJ & 1.04mJ & NR & NR & \textbf{0.29mJ - 5.97mJ}  \\
        \midrule
        Neuron Model     & LIF & IF & LIF/Izk. & LIF & NR & IF \\
        \# Neuron        & 384  &  400 & 256 & 2048 & 128 & 2000 -- 9000 \\
        \# Synapses      & 176800  & 230400 & 64000 & 2M & NR & 1.6M -- 7.2M \\
        Weight Precision & 9b fixed  & 1b & 4b & 8b & 8b &  1b \\
        Encoding         & temporal & rate  & rate \& rank \tnote{$\diamondsuit$} & temporal & TTFS & rank-1 \\
        \midrule
        \multicolumn{7}{c}{Learning specs}\\
        \midrule
        Online  & yes  & yes & yes & yes & yes & yes\\
        On-chip & yes  &  yes & yes & yes & yes & yes\\
        Rule    & STDP var. & spike cnt & SDSP & addSTDP & DRTP & BS-STDP \\
        \midrule
        \multicolumn{7}{c}{Hardware specs} \\
        \midrule
        ASIC Node        & 28nm & 65nm & 28nm & 28nm & 28nm & 22nm \\
        Clock Frequency  & 333MHz & 384MHz & 75MHz & 500MHz & 150MHz & 500MHz  \\ 
        Voltage          & 0.9V & 1.2V & 0.55V & 0.81V & 0.6V  & 0.8V \\
        Area ($mm^2$)    & 1 & 0.39 & 0.086 & 6.22 & 0.26 & 0.232 -- 0.757 \\ 
        Norm. area\tnote{$\S$} ($mm^2$) &  0.617 & 0.044 & 0.053 & 3.83 & 0.16 & 0.232 -- 0.757  \\
        \midrule
        \multicolumn{7}{c}{Energy} \\
        \midrule
        Learning  & 660nJ & 2630nJ & 105nJ       & NR     & NR & 148.4nJ -- 663.5nJ\\
        \hspace{1em}per SOP   & NR    & 1.42pJ & NR          & 4.99pJ & NR    & 1.5pJ\\
        Inference & 500nJ & 310nJ  & 15nJ--404nJ & NR     & 313nJ    & 147.2nJ -- 662.3nJ\\
        \hspace{1em} per SOP   & NR    & 0.26pJ & 12.7pJ      & 1.28pJ & NR    & 0.09pJ\\
        \midrule
        \multicolumn{7}{c}{Norm. Energy\tnote{$\dagger$}} \\
        \midrule
        Learning  & 409.7nJ & 395.62nJ & 174nJ             & NR     & NR & 148.4nJ -- 663.5nJ\\
        \hspace{1em} per SOP  & NR      & 0.21pJ   & NR                & 3.82pJ & NR      & 1.5pJ\\
       Inference & 310.4nJ & 46.6nJ   & 24.93nJ--671.58nJ & NR     & 437.2nJ      & 147.2nJ -- 662.3nJ\\
        \hspace{1em} per SOP   & NR      & 0.03pJ   & 21.11pJ           & 0.98pJ & NR      & 0.09pJ\\
        \midrule
        \multicolumn{7}{c}{Throughput (fps)} \\
        \midrule
       Learning & 211.77k & NR & NR & NR & NR & 236.4K -- 54.8K \\
       Inference & 277.78k & NR & NR & NR & 8.5K & 248.1K -- 55.4K \\
        \midrule
        \multicolumn{7}{c}{Learning energy overhead for inference \tnote{$\nabla$}} \\
      \midrule  
        & 0.32 & 7.5 & 7 \& 0.25 \tnote{$\diamondsuit$} & 2.89 & NR & $<$\textbf{0.01} \\
        \bottomrule
\end{tabular*}
            
    \begin{tablenotes}[para,flushleft,small]
        \item NR: Not Reported\, \item addSTDP: Additive STDP\,
        \item BS-STDP: Stochastic Binary STDP \, \\
        \item DTRP: Direct Random Target Projection - a modified back-propagation version, suitable for online and local learning \\
        \item[$\triangleq$] Number of weight updates is inferred from the reported number of neurons allowed to learn each input sample multiplied by the number of training samples used to reach accuracy
        \item~\cite{tcsi2022,tbcas2109} report area/energy results post-layout and they do not include I/O area/energy cost; \cite{odin,tcsii2023,spoon} report area/energy results post-tapeout. \\
        \item[$\ddag$] Reported accuracy for one epoch is 92.8\%,  and after 100 epochs it reaches 95.3\%. \\
        \item[$\S$] $ \text{Area} \times  (22 / \text{Node} )^2$\\ 
        \item[$\dagger$] $ \text{Energy} \times (22/ \text{Node}) \times (0.8 / \text{voltage})^2 $ \\
        \item[$\diamondsuit$] \cite{odin} supports both rate coding  and temporal, rank-order coding for input spikes. Reported inference energy and learning overhead is for rank and rate coding, respectively. Reported learning energy excludes inference cost.
        \item[$\nabla$] Energy overhead computed as $( \text{learning energy} - \text{inference energy})/\text{inference energy}$
    \end{tablenotes}
\end{threeparttable}
\end{table*}

\begin{figure*}
    \begin{subfigure}{.5\textwidth}
        \centering
        \includegraphics[width=\linewidth]{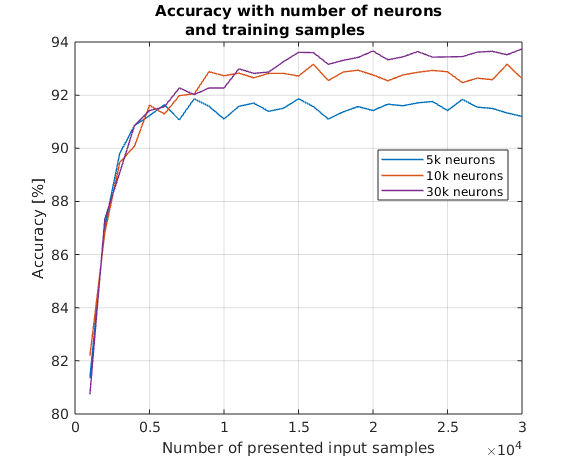}
        \caption{}
        \label{fig:mnist_acc_samples}
    \end{subfigure}
    \begin{subfigure}{.5\textwidth}
        \centering
        \includegraphics[width=\linewidth]{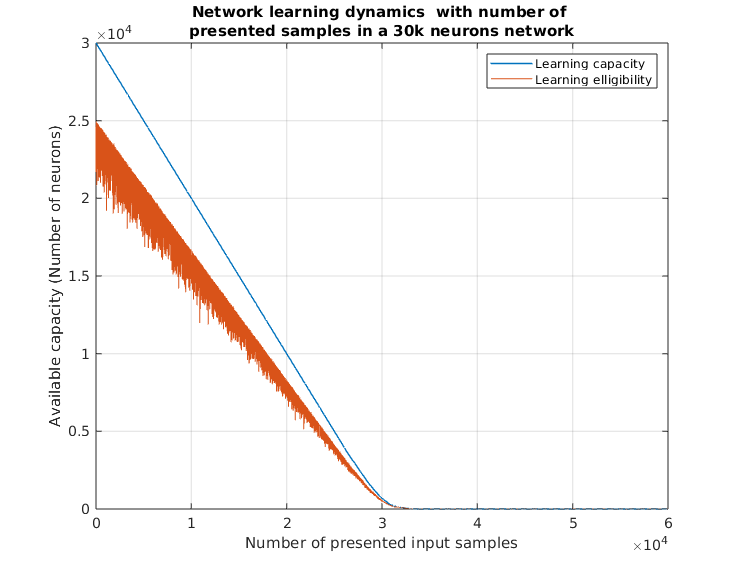}
        \caption{}
        \label{fig:mnist_netw_cap}
    \end{subfigure}
    \caption{Online learning performance on the MNIST digit classification task. (a) Accuracy as a function of data samples presentation for three IF layer sizes. (b) Dynamics of a 30k network learning capacity and eligibility, as a function of presented samples}
    \label{fig:mnist_algo_1}
\end{figure*}

\begin{figure*}
    \begin{subfigure}{.5\textwidth}
        \centering
        \includegraphics[width=\linewidth]{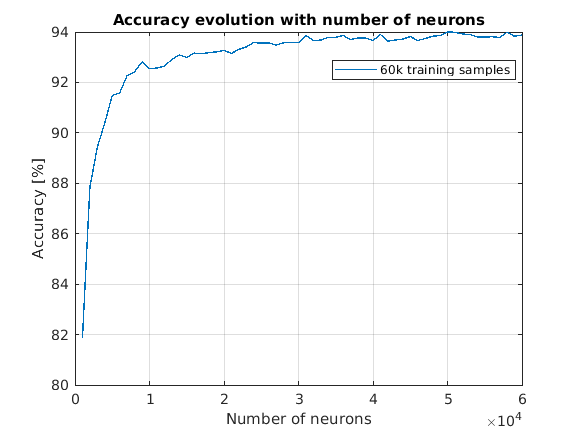}
        \caption{}
        \label{fig:mnist_acc_neu}
    \end{subfigure}
    \begin{subfigure}{.5\textwidth}
        \centering
        \includegraphics[width=\linewidth]{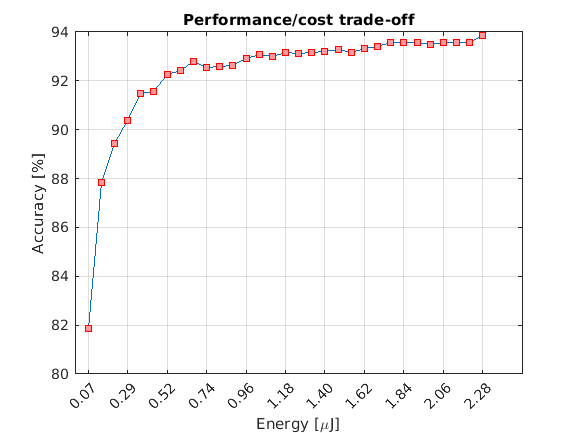}
        \caption{}
        \label{fig:mnist_acc_energy}
    \end{subfigure}
    \caption{Online learning performance on the MNIST digit classification task.k. (a)  Accuracy as a function of the IF layer size for 60K samples. (b) Accuracy versus energy while scaling the IF layer from 1K neurons to 30K neurons.}
    \label{fig:mnist_algo_2}
\end{figure*}

\subsection{Algorithm performance and hardware results benchmarking --multiclass classification task}


Fig. \ref{fig:mnist_algo_1} and Fig. \ref{fig:mnist_algo_2} show the online learning algorithm performance for the MNIST dataset. Fig. \ref{fig:mnist_acc_samples} demonstrates how accuracy continuously increases with more training samples, but also with a higher initial network capacity (i.e., number of IF neurons in the trainable layer; initially, the network learning capacity is $N$, but this number decreases with more learned samples). The plot also shows how, once the number of presented training inputs approaches the network's initial capacity, accuracy saturates. This behaviour is likewise exhibited in Fig. \ref{fig:mnist_netw_cap}, which shows that, with more training samples being presented, less neurons are eligible to learn (i.e., network learning eligibility refers to the number of neurons that pass their learning threshold for a presented input sample), and as a result, the network's learning capacity decreases. While this outcome is expected in all neural networks, it is worth noting that in this work, the
learning occurs continuously, on-device. Moreover, here, the steep and monotonic curve behaviour is due to our choice of learning parameters for fast learning, that forces individual neurons to "lock" to a single input pattern, and there is no learning threshold decay to allow them to forget or re-learn. Concurrently, as a result of stochastic learning, there is no overfitting, and the network can generalize well on previously unseen input samples. The latter result also indicates that, since we only learn $W$ random connections, our network behaves like an associative memory, e.g., it can recall patterns that are degraded or only partially resemble learned inputs. Finally, Fig. \ref{fig:mnist_acc_neu} and Fig. \ref{fig:mnist_acc_energy} show how increasing the network capacity improves accuracy, on the one hand, but also how it affects inference energy cost, on the other hand.


In the hardware measurements for this task, we did not use parallelization of the IF neuron processing, and therefore the latency reported is measured as $(Spike\_Vec\_Gen\_Ovh + N + 1) \times clk\_period$. Thus, for e.g., for a 10 $ns$ clock period (FPGA) and $2000$ neurons, and taking into account the spike encoder overhead, the latency is $(14 + 2000 + 1) \times 10 = 20150ns$ and the throughput is $10^9 / 20150 = 49627$ inferences (frames\footnote{One inference is one input digit frame for these measurements}) per second.

For the ASIC instantiation of EON-1, area and energy measurements exclude I/O cost and include memory area and access cost. Online learning energy consumption is computed as the total energy cost for performing one inference, in order to find the neuron that reaches its learning threshold, followed by the energy cost of the neuron updating its weights. Based on these energy costs, we compute the learning overhead as $(E_{learn} - E_{inf})/E_{inf}$. 

Tables \ref{tab:fpga_benchm} and \ref{tab:asic_benchm} 
compare, based on various metrics,
EON-1 against other FPGA and ASIC-based solutions in the recent literature, on the MNIST benchmark.
From the comparison with FPGA solutions, EON-1 has the highest throughput. We mainly attribute this result to using binary weights and avoiding multiplications throughout the entire design, such that only bitwise operations and additions are performed, thus allowing for more operations to be scheduled within one clock cycle. Moreover, except from the work in \cite{tcsii2021}, which does not embed learning on-chip, EON-1 also stands out in terms of resources utilization, thus highlighting the hardware-friendliness of our inference and learning approach in this work.

For ASIC comparison, we use the method in \cite{rabaey} to normalize the reported area and energy of these solutions to our 22nm node. For EON-1, we report results for reaching accuracies between 87.8\% -- 92.8\% for a closer comparison to other work. What stands out from these comparisons is that EON-1, while competitive in performance with the state-of-the-art:
\begin{enumerate}[(a)]
    \item  it has by more than an order of magnitude lower energy overhead for supporting on-device learning (overhead which includes the cost of spike encoding at the input and readout at the output);
    \item  for converging to the reported accuracy, it is the most energy efficient; 
    \item it achieves one of the highest throughputs;
    \item  it is the only one that supports very fast (worst case latency is $D^2 \times clk\_period$) online learning with binary and stochastic STDP, both in FPGA and ASIC.
\end{enumerate}


\subsection{Streaming data processing -- binary classification}

\begin{figure}
        \centering
        \includegraphics[width=\linewidth]{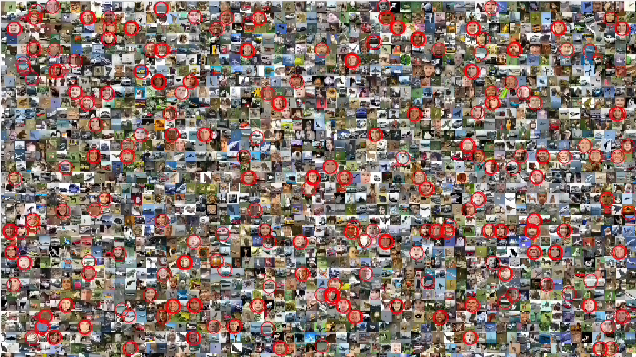}
        \caption{One quarter of the collage of many small faces and non-faces images, randomly placed in one UHD frame. The full collage frame contains 798 faces.}
        \label{fig:uhd_img}
\end{figure}

One of the main motivations of this exploration has been to see if EON-1's fast learning and inference can sustain the processing of continuous streaming data from a high-resolution image sensor. In this experiment we aimed to test just that. We combined face images from the UTK Face Dataset~\cite{utkface} (which we have downscaled to $32 \times 32$) with non-face images from the CIFAR-10~\cite{cifar} dataset (where we keep the initial resolution of $32 \times 32$) to create high-resolution (UHD) collage frames such as the one shown in Fig. \ref{fig:uhd_img}. The task here is to detect and adapt to faces in the UHD image. However, this image cannot be processed in one go as a single patch residing in memory, but rather a rolling window needs to parse the image with a stride and load a patch-at-a-time to the EON-1 engine, similar to 2D convolution with a kernel. Thus, the challenge here is to see if the speed of processing of these images with EON-1 is acceptable for real-time video inference and online adaptation.

Consequently, we defined two types of performance measurement tests: in the first one, we measure accuracy performance on a balanced testset of 10000 samples of faces and CIFAR-10 samples, different from the trainset. For the second test, we measure the network's recall per UHD frame, i.e., we are interested in how many faces we can identify with EON-1 in the entire frame. Note that, in this situation, the field of vision (FoV) -- which we denote by one patch -- often contains both faces and non-faces (due to using a sliding window with stride 1), introducing a degree of position variance to the tested inputs. Following, we describe the two experiments we have run for this task.

The first experiment aimed to evaluate the performance of a fully pre-trained network (i.e., the network learning capacity was saturated by presenting a number of samples equal to the network size). We initially trained a network of 400 neurons, each with 3136 synapses ($4 \times 28 \times 28$), using only faces. This network reached 95.7\% accuracy on the previously mentioned balanced testset. Using the pre-trained weights, we then evaluated the network's inference performance on the UHD frame. For this case, 720 out of 798 faces (i.e., recall is 90.2\%) randomly placed in the UHD frame were correctly identified. 

\begin{figure}
    \centering
    \includegraphics[width=\linewidth]{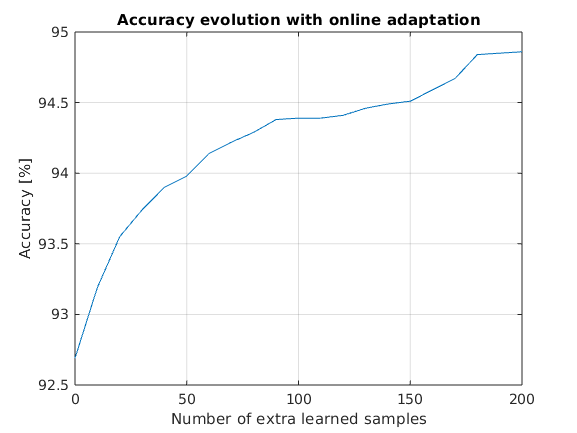}     
    \caption{Online on device adaptation on the face detection task: accuracy as a function of online adaptation to more data samples in a pre-trained network, on the UHD frame processing with a rolling window. }
   \label{fig:faces_acc_adapt}
\end{figure}

For the second experiment, the goal was to evaluate online adaptation on input patches extracted from the UHD frame, starting from a network that was pre-trained on a number of inputs samples that is much smaller than the initial network learning capacity. This use case can occur, for instance, when not enough training data is available. The adaptation is enabled in a self-supervised manner, such that learning is triggered whenever the network fires (i.e., it indicates a face), until it reaches it's learning capacity. In Fig. \ref{fig:faces_acc_adapt} we depict how, in a network with an initial capacity of 400 neurons, that we pre-trained on 200 samples, the accuracy on the balanced testset slowly increases from 92.69\% before adaptation to 94.86\% after self-supervised learning of more samples. Before adaptation, the recall on the UHD frame is 72\% (587 faces are identified) and after exposure to new data from the UHD frame patches, the recall reaches 85\% (680 faces are identified). The latter result shows a lower recall than for the fully pre-trained network of the previous test, however, this is an expected result, since the adaptation occurs on patches that contain position variance, as a result of the FoV extraction. Nevertheless, despite it's simplicity, this experiment shows that it is possible to adapt on-the-fly to new inputs, when few data samples are available for training. 

Finally, we report hardware performance metrics for this task in Table \ref{tab:uhd_speed}. Since we use a stride of 1 pixels, parsing one UHD frame involves processing 8 million patches, for which the projected energy consumption in the GF22nm ASIC node is 1.85J. Although this use-case is not low-power, the goal here was to explore the maximum attainable speed in processing UHD frames and while having learning active.
We therefore process the full high-resolution frame, without any downscaling, which is a typical pre-processing step for real-time video processing with neural networks. At full UHD resolution, we were able to process 12 FPS in FPGA, at 100MHz, by using a parallelization factor of $P = 400$ (see Fig. \ref{fig:multiple_match_units}; this limit is imposed by the maximum achievable bandwidth on the FPGA platform used in this work). This is about half the baseline framerate for real-time motion video (25fps) and 1/5 of high-quality UHD video (60fps). Thus, either downscaling the UHD frame, before processing it using EON-1, or operating the circuit at the ASIC clock frequency (500MHz) would make real-time UHD video processing indeed possible.


\begin{threeparttable}[htb]

    \footnotesize
    \caption{Face detection task: performance metrics for processing UHD frames with EON-1}
    \label{tab:uhd_speed}
     \begin{tabular}{l|c|c | c }
     \hline
            Resolution & Latency/frame &  FPS    & Energy/frame \\
            & (100MHz) & (100MHz) &  (in ASIC) \\
            \hline
            2160×3840 \tnote{1} & 84ms  & 12  &  1.85J  \\
            1080x1920 \tnote{2} & 20 ms & 50  &  0.45J   \\
            1280x720 \tnote{3}  & 9ms   & 111 & 0.19J     \\
            \hline
     \end{tabular}

        \begin{tablenotes}[para,flushleft,small]
       \item[1] UHD resolution
       \item[2] Full HD resolution
       \item[3] HD resolution 
    \end{tablenotes}
\end{threeparttable}

\vspace{1em}

\section{Conclusions}

 In this paper, we introduced EON-1, an energy- and latency-efficient custom edge processor for Spiking and Convolutional Neural Networks, that uses 1-bit synaptic weights and 1-spike per neuron. EON-1 embeds a very fast on-device online learning algorithm (amortizable for few-shot learning) founded on binary stochastic STDP, which, benchmarked in an ASIC node, achieves less than 1\% energy overhead for on-device learning (relative to the inference). To our knowledge, our solution incurs the least energy overhead for learning on-device, compared to state-of-the-art solutions, showing a better efficiency by at least a factor of 10x.  We report that EON-1 consumes 0.29mJ to 5.97mJ for converging to accuracies between 87.8\% - 92.8\% on the MNIST classification task, from randomly initialized weights for both inference and on-device learning. With an energy consumption of 148.4nJ - 663.5nJ per processed sample, only 1.2nJ/sample is used for learning. We extend our solution to a practical use-case for real-time processing of UHD videos, consuming 1.85J for each UHD frame.


\vspace{12pt}
\color{red}

\end{document}